\documentclass{article}
\usepackage{spconf,amsmath,multirow,graphicx}
\usepackage{amssymb}

\title{When Training-Free NAS Meets Vision
Transformers: A Neural Tangent Kernel Perspective}
\name{Qiqi Zhou$^{1,2}$, Yichen Zhu$^{1,*}$\thanks{* Corresponding Author.}}
\address{$^1$Shanghai University of Electric Power, College of Mathematics and Physics \\ $^2$Midea Group}
\begin{document}
%\ninept
%
\maketitle
\begin{abstract}
This paper investigates the Neural Tangent Kernel (NTK) to search vision transformers without training. In contrast with the previous observation that NTK-based metrics can effectively predict CNN’s performance at initialization, we empirically show their inefficacy in the ViT search space. We hypothesize that the fundamental feature learning preference within ViT contributes to the ineffectiveness of applying NTK to NAS for ViT. We both theoretically and empirically validate that NTK essentially estimates the ability of neural networks that learn low-frequency signals, completely ignoring the impact of high-frequency signals in feature learning. To address this limitation, we propose a new method called ViNTK that generalizes the standard NTK to the high-frequency domain by integrating the Fourier features from inputs. Experiments with multiple ViT search spaces on image classification and semantic segmentation tasks show that our method can significantly speed up search costs over prior state-of-the-art NAS for ViT while maintaining similar performance on searched architectures.
\end{abstract}
\section{Introduction}
Vision Transformer (ViT) is increasingly matching CNNs in various vision tasks in various domains such as VQA and robotics~\cite{zhu2024llava, liu2023query, wen2024object, zhu2024language}. Prior studies have presented many ways to accelerate vision transformer, i.e., distillation~\cite{zhu2021student,zhu2022teach,zhu2023scalekd}, dynamic networks~\cite{zhou2023make} Neural architecture search (NAS). The latter aiming for hands-off architectural design, has successfully influenced CNNs across tasks. Yet, traditional training-based NAS methods are resource-draining. There's a rising interest in training-free NAS, which assesses untrained architectures using performance predictors. Given ViT's complex structure, including elements like multi-head self-attention (MSAs) and feed-forward networks (FFN), the reliability of training-free metrics tailored for CNNs in the ViT context is uncharted. 
\begin{figure}[t]
     \centering
     \includegraphics[width=0.5\textwidth]{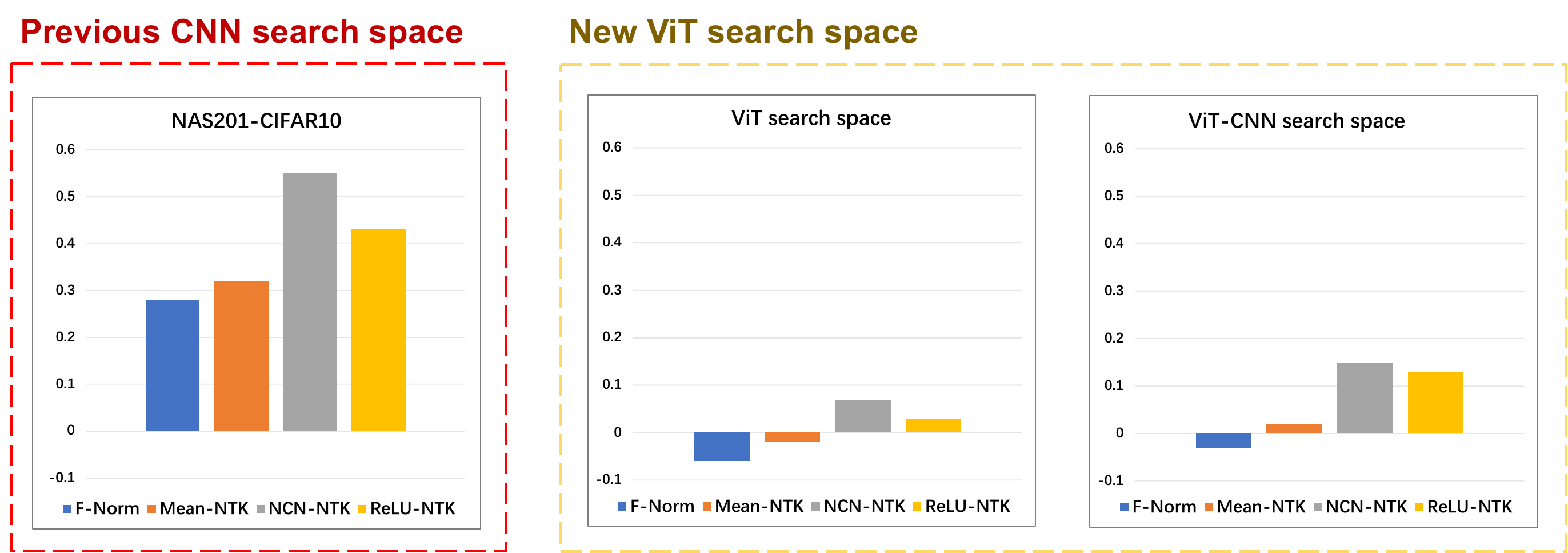}
     \caption{The correlation analysis of NTK-based metrics on CNN and ViT search spaces. \textbf{Left:} Kendall-Tau value on NAS-Bench-201~\cite{dong2020bench}. \textbf{Right:} Kendall-Tau value on two ViT search spaces.}
     \label{fig:ntk_kendalltau}
\end{figure}
The shortcomings of NTK-based metrics for ViT are intriguing. We hypothesize that this is due to ViT's mixed feature learning methods. Though its self-attention mechanism functions as a low-pass filter, ViT still leans on high-frequency signals for discerning images. This duality can skew metrics. Our study shows that NTK primarily reflects networks' low-frequency learning prowess. Tests focusing on ViT's MSAs affirm that NTK can adeptly pinpoint potent ViT architectures, reinforcing MSAs' role as established low-pass filters.

This makes us wonder what causes the inefficacy of NTK-based metrics on ViT search spaces. We posit that this observation is attributed to the conflict in feature learning of ViT. Recall that the self-attention mechanism acts as a low-pass filter. The ViT still relies on high-frequency signals to discriminate images. As such, the conflict feature learning capacity between different components confuses the metrics. We give a theoretical analysis to understand this phenomenon by proving an approximation error of NTK's predictive ability on deep neural networks (DNNs) under mild conditions. Specifically, our analysis shows that NTK-based metrics are strongly correlated to the dimension of low-frequency signals. In other words, NTK can reasonably approximate the strength of low-frequency signals learned by neural networks. As proof of concept, we again assess the NTK-based metrics on ViT search space, \textit{searching multi-head self-attention (MSAs) only}. Surprisingly, the experimental results show that NTK can find promising ViT when only MSAs, provable low-pass filters, are involved in the search process.

While current analyses of NTK on ViT seem adequate, they reveal that NTK-based training-free metrics aren't ideal for finding top-tier ViT. We introduce the Fourier Neural Tangent Kernel, transforming NTK metrics using Fourier feature mapping. This upgrade enhances NTK's capacity to predict high-frequency signals, mitigating the conflict between MSAs and other high-frequency modules.

\section{Analysis of Using NTK-Based Metrics to Search ViT}

\subsection{Preliminary} 
Suppose there are $D$ training points denoted by $\{(x, y)_{d=1}^{D}\}$, where input $X = (x_{1}, \dots, x_{D}) \in \mathbb{R}^{n_{0} \times D}$, and label $Y = (y_{1}, y_{D}) \in \mathbb{R}^{n_{L} \times D}$. We consider either a fully connected network or a convolutional neural network. Since our analysis does not involve these two networks, We omit the detailed descriptions for simplicity.

A large body of recent works~\cite{bietti2019inductive} has shown that when training deep networks in a certain over-parameterized regime, the dynamic of gradient descent behaves like those of linear models on non-linear features determined at initialization. With sufficient over-parameterization, the optimization dynamics converge to a certain gram matrix in the reproducing kernel Hilbert space (RKHS) known as the neural tangent kernel $\Theta_{ntk}(x,x'):= \langle \nabla_{\theta}f(x,\theta), \nabla_{\theta}f(x',\theta))\rangle$. Where function $f$ is the output of the network at training time, i.e., $f(x, \theta) = h_{t}^{L}(\theta, X) \in \mathbb{R}^{D \times n_{L}}$, and $\nabla_{\theta}f(\theta, X) = vec([\nabla_{\theta}f(x, \theta)]_{x \in X}) \in \mathbb{R}^{D_{n_{L}}}$. 

In this limit and under some assumptions, one can show that weights barely move, and the kernel remains fixed during training. Such property motivates recent training-free NAS to directly calculate some NTK-based metrics for random initialized CNNs and use the metrics as a performance proxy. 

\subsection{Rank Correlation Analysis}
\label{sec:rank_cor}
In this section, we evaluate how reliable NTK-based metrics are for training-free NAS. We compare with four variants of NTK-based metrics and refer readers to more detailed descriptions of their papers. These methods include F-Norm NTK~\cite{xu2021knas}, Mean NTK~\cite{xu2021knas}, NCN NTK~\cite{chen2021searching}, and ReLU NTK~\cite{cho2009kernel}. We conduct experiments on a pure ViT search space and a hybrid ViT-CNN search space. For each ViT-based search space, we randomly sample 100 architectures. All of them are trained to convergence, and the corresponding metrics are reported. Finally, we briefly introduce two ViT-based search spaces, where the comprehensive description of the listed search space is presented in the Appendix.

\noindent
\textbf{ViT search space.} We construct a simple, pure ViT search space that is composed of kernel size (used to split image patches) and expansion number for feed-forward networks. We follow the Twins~\cite{chu2021twins} to build a four-stage ViT, a down-sample module at the top of each stage to reduce the spatial dimension of feature maps gradually. 

\noindent
\textbf{ViT-CNN hybrid search space.} We follow the NASViT~\cite{gong2022nasvit} to construct a ViT-CNN hybrid search space to verify our claim. The search space consists of two CNN blocks followed by two transformer blocks. A downsampling module is inserted prior to each building block. We search for the kernel size and SE module in the CNN block, as well as the expansion number in the Transformer block.

In Figure~\ref{fig:ntk_kendalltau}, we report the Kendall-tau correlations $\mathbf{\tau}$ between these metrics and the model’s ground-truth accuracies. On both ViT search spaces, the Kendall-Tau correlation on all four metrics is low: on the pure ViT search space, the highest Kendall-tau is 0.07, and the lowest Kendall-Tau is -0.05. Similar results are obtained on hybrid ViT-CNN search space, with a slightly better highest Tau value $\tau = 0.14$. These are counter-intuitive results since NTK is very effective on NAS-Bench-201~\cite{dong2020bench}. For instance, NCN-NTK achieves over $\tau$ value over 0.5 on NAS-Bench-201; and even the worse metric, F-Norm, obtains $\tau = 0.28$. Based on our observation, it seems that the magic of NTK-based metrics is somehow gone in the ViT search space. The following section sheds light on this question.

\subsection{Why NTK Fails to Find Good ViT?}\label{sec:whyfail}
In this section, we provide a theoretical justification to answer why NTK fails to find good ViT. First, we provide an approximation error for NTK on deep neural networks (DNNs), which is extended from Ghorban et al.~\cite{ghorbani2020neural}. We show that NTK is a better approximator for DNN that relies on low-frequency features. We further provide a proof of concept to show our justification can indeed explain such phenomena. 

We start our analysis by setting up a simple proxy for Fourier decomposition, the decomposition of covariate vector $\mathbf{x}$, where its low-dimensional dominant component and high-dimensional dominant component represent low-frequency and high-frequency features, respectively. Then, we introduce the spiked covariates model that approximates low-dimensional covariates. That is, we not only expect the target function $f^{*}$ to depend predominantly on the low-frequency components of image $\mathbf{x}$, but the image $\mathbf{x}$ itself to have most of its spectrum concentrated on low-frequency components. As such, our expected target function would be the neural network that can approximate the low-frequency features. Specifically, we consider $\mathbf{x} = \mathbf{U}\mathbf{z}_{1} + \mathbf{U}^{\perp}\mathbf{z}_{2}$, $\mathbf{U}^{\perp} \in \mathbb{R}^{d \times d_{0}}$, $\mathbf{U} \in \mathbb{R}^{d \times (d-d_{0})}$, $[\mathbf{U}|\mathbf{U}^{\perp}] \in \mathbb{R}^{d \times d_{0}}$ is an orthogonal matrix. We assume that $\mathbf{z}_{1}$ and $\mathbf{z}_{2}$ are uniformly distributed over the sphere $\mathbf{z}_{1} \sim U(S^{d_{0}-1}(r_{1}\sqrt{d_{0}}))$, $\mathbf{z}_{2} \sim U(S^{d-d_{0}-1}(r_{1}\sqrt{d-d_{0}}))$, and $r_{1} \geq r_{2}$. The $d_{0}$ is signal dimension, $d$ is ambient dimension, and $\frac{r_{1}}{r_{2}}$ is covariate signal-to-noise ratio. We can have the next theorem that characterizes the asymptotics of the approximation error for NTK on NN:

\noindent
\textbf{Theorem 1.} (Approximation of NTK for NN) \textit{Without loss of generality, we assume the target function $f_{*}(\mathbf{x})=\sigma(\mathbf{U}^{T}\mathbf{x})$. Assume that $\sigma \in \mathcal{C}^{\infty}(\mathbf{R}), \sigma^{k}(x)^2 \leq c_{0,k}e^{c_{1,k}\frac{x^2}{2}}$, for $\forall k, \forall x \in \mathcal{R}, c_{0,k} > 0$ and $c_{1,k} < 1$, then we have}
\begin{equation}
\begin{split}
|\mathbf{R}_{NN,N}(f_{*};\mathbf{W})- \mathbf{R}_{NTK,N}(f_{*};\mathbf{W})| \leq \\ o_{d}(1)\cdot||P_{>l+1}f_{*}||_{L^2}(1-||f_{*}||^{2}_{L^2}+\tau^2)
\end{split}
\end{equation}
The $P_{>k} = I - P_{\leq k}$ and $P_{\leq k}: L^{2} \rightarrow L^{2}$ is a mapping function to the space of degree k orthogonal polynomials. The $\mathbf{R}_{NN}$ and $\mathbf{R}_{NTK}$ denote the prediction error of neural networks and NTK with respect to the data. Remarkably, the above theorem shows that the effective signal dimension controls the sample complexity N for NTK in approximating an NN to learn a degree l polynomial. Based on our initial setting, the low-frequency components effectively contribute to the capacity of NTK on the ability of performance estimation of NN. Theorem 1 explains that NTK can achieve remarkably better approximation on NN because ViT extracts low-frequency signals to recognize the object.

\begin{figure}[t]
    % \begin{minipage}{1.0\linewidth}
    \centering
    \includegraphics[width=0.45\textwidth]{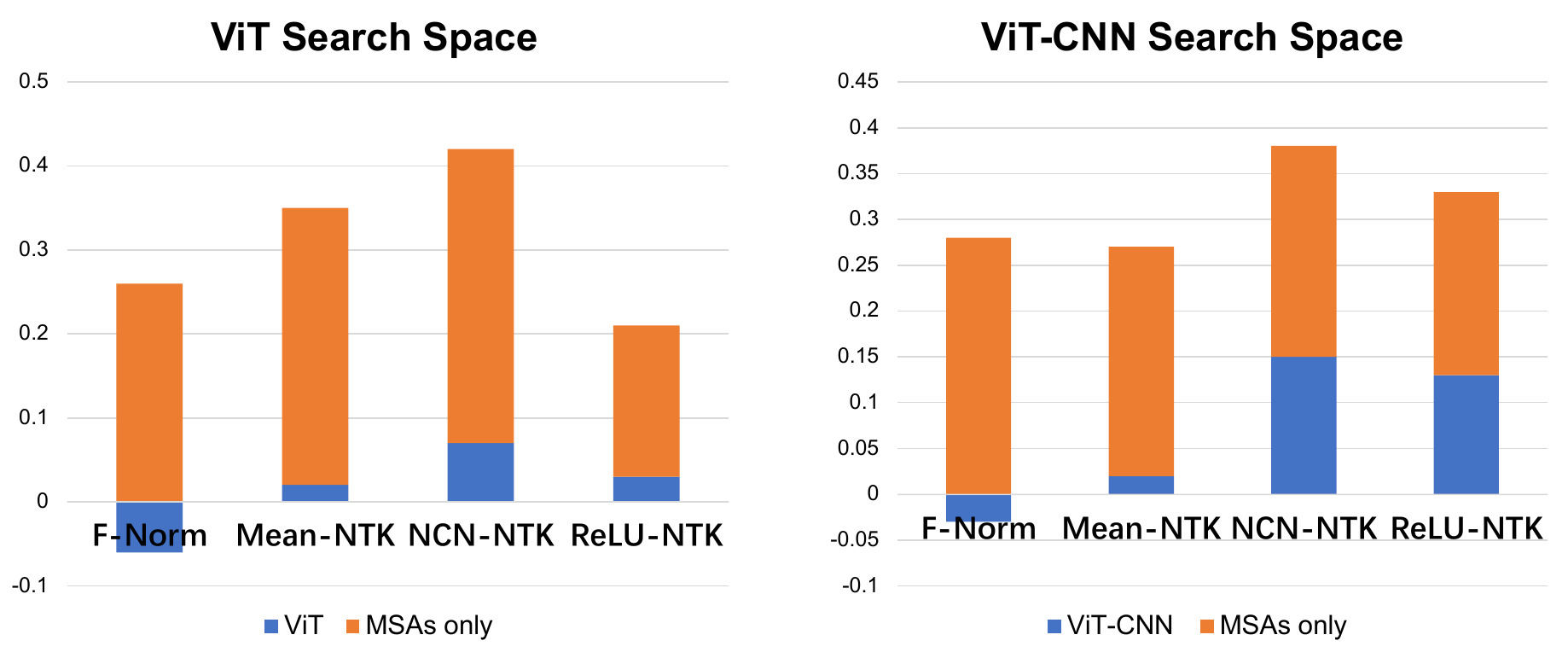}\\
    % \end{minipage}
    % \vspace{-3 mm}
      \caption{The Kendall-Tau correlation for MSAs only in two ViT search spaces. The value of $\tau$ clearly improves when only MSAs are involved in the search space.}\label{fig:msaonly}
\end{figure}

\begin{table}[t]
  \centering
  \caption{Image Classification on ImageNet-1k with AutoFormer~\cite{chen2021autoformer} search space.}
  \label{tbl:autoformer}
  \renewcommand*\arraystretch{1.1}
  \resizebox{0.45\textwidth}{!}{\begin{tabular}{c|c|c|c|c|c}
    \hline
    Method & Param (M) &  FLOPs (B) & Top-1 (\%) & Top-5 (\%) & Search Cost\\
    \hline
    DeiT-T~\cite{touvron2021training} & 5.7 & 1.2 & 72.2 & 91.1 & - \\

    ViTAS-C~\cite{su2021vision} & 5.6 & 1.3 & 74.7 & 91.6 & 32 \\
    AutoFormer-T~\cite{chen2021autoformer} & 5.7 & 1.4 & 74.7 & 92.6 & 24 \\
    TF-TAS-Ti~\cite{tftas} & 5.9 & 1.4 & 75.3 & 92.8 & 0.5 \\
    \textbf{ViNTK-T} &  \textbf{5.8} & \textbf{1.3} & \textbf{75.6} & \textbf{93.1} & 0.8 \\
    \hline
    DeiT-S~\cite{touvron2021training} & 22.1 & 4.7 & 79.9 & 95.0 & - \\
    Swin-T~\cite{liu2021swin} & 29.0 & 4.5 & 81.3 & - &- \\

    ViTAS-F~\cite{su2021vision} & 27.6 & 6.0 & 80.5 & 95.1 & 32 \\
    AutoFormer-S~\cite{chen2021autoformer} & 22.9 & 5.1 & 81.7 & 95.7 & 24 \\
    TF-TAS-S~\cite{tftas} & 22.8 & 5.0 & 81.9 & 95.8 & 0.5\\
    \textbf{ViNTK-S} & \textbf{22.7} & \textbf{4.9} & \textbf{82.3} & \textbf{96.2} & 0.8 \\
    \hline
    DeiT-B~\cite{touvron2021training} & 86.0 & 18.0 & 81.8 & 95.6 & - \\
    Swin-B~\cite{liu2021swin} & 88.0 & 15.4 & 83.5 & - & - \\

    AutoFormer-B~\cite{chen2021autoformer} & 54.0 & 11.0 & 82.4 & 95.7 & 24 \\
    TF-TAS-B~\cite{tftas} & 54.0 & 12.0 & 82.2 & 95.6 & 0.5\\
    \textbf{ViNTK-B} & \textbf{53.3} & \textbf{10.8} & \textbf{82.3} & \textbf{95.6} & 0.8\\
    \textbf{ViNTK-L} & \textbf{53.9} & \textbf{12.0} & \textbf{83.0} & \textbf{96.3} & 0.8\\
    \hline
    \hline
  \end{tabular}}
\end{table}

\noindent
\textbf{Proof of concept.} Our theory suggests that NTK effectively approximates networks learning low-frequency signals. Thus, we divide the search space: MSAs represent low-frequency components, while convolutional layers, patch embedding, and feed-forward networks represent high-frequency. Only MSAs are available for search, ensuring other architectural capabilities remain constant. This means NTK-based metrics should mainly reflect the strength of low-frequency signals.

Figure~\ref{fig:msaonly} showcases the correlation between NTK metrics and accuracy when only MSAs are considered. The correlation is significantly stronger than in a full search space. Specifically, the correlation values in pure ViT and hybrid ViT-CNN spaces increase by 0.28 and 0.24, respectively, confirming our theoretical insights and highlighting potential shortcomings in NTK-based metrics.

\section{Fourier Neural Tangent Kernel}
Drawing inspiration from the Fourier features, we propose to align the Fourier features with NTK. Specifically, consider optimizing a parametric model with some bounded loss $l(y, \hat{y})$, where $\hat{y}$ is the prediction of the models on the training dataset. We can rewrite the NTK by taking the eigendecomposition of NTK matrix $\Theta_{ntk}$ into $\Theta_{ntk} = QAQ^{T}$, where $Q$ is an orthogonal matrix and $A$ is a diagonal matrix whose entries are the eigenvalues $\lambda_{i} \leq 0$ of $\Theta_{ntk}$, given $\Theta_{ntk}$ is a symmetric positive semi-definite matrix. As a result, we reformulate the NTK matrix:
\begin{equation}
    \Theta_{ntk} = -e^{-\eta At}Q^{T}y
\end{equation}
The above equation represents that in the eigenbasis of the NTK, the convergence rate of $i^{th}$ component of NTK will decay approximately exponentially at the rate of $\eta \lambda$~\cite{ronen2019convergence}. Consequently, the convergence of high-frequency signals is slower than low-frequency signals. We would like to fix this issue by balancing the convergence rate on the Fourier spectrum. 

We leverage random Fourier features~\cite{rahimi2008weighted}, introduced to approximate an arbitrary stationary kernel function via Bochner's theorem. We use their method to feature input on the Fourier domain and combine it with NTK. Specifically, a Fourier feature mapping function $\gamma$ transforms input to the surface of a higher dimensional hyper-sphere and obtains the Fourier kernel function as
\begin{equation}
    \Theta_{fourier}(x,x') = \gamma(x)^{T}\gamma(x') = \sum^{m}_{j=1}\alpha^{2}_{j}cos(2\pi b^{T}_{j}(x-x'))
\end{equation}
where $\gamma(\cdot)$ is a set of sinusoids. Combined with NTK, we now have
\begin{equation}
    \Theta_{vintk} := \Theta_{ntk}(x,x') \circ \Theta_{fourier}(x,x')
\end{equation}
In practice, the $b_{j} = j$ is set as full Fourier basis, $\alpha_{j} = \frac{1}{j_{p}}$ and $p=2$ for $j=1, \dots, \frac{n}{2}$. For NTK, we directly adopt Mean-NTK. The extra computational cost of calculating ViNTK is negligible, making our method ultra-fast at the search stage.

\begin{figure}[t]
     \centering
     \includegraphics[width=0.22\textwidth]{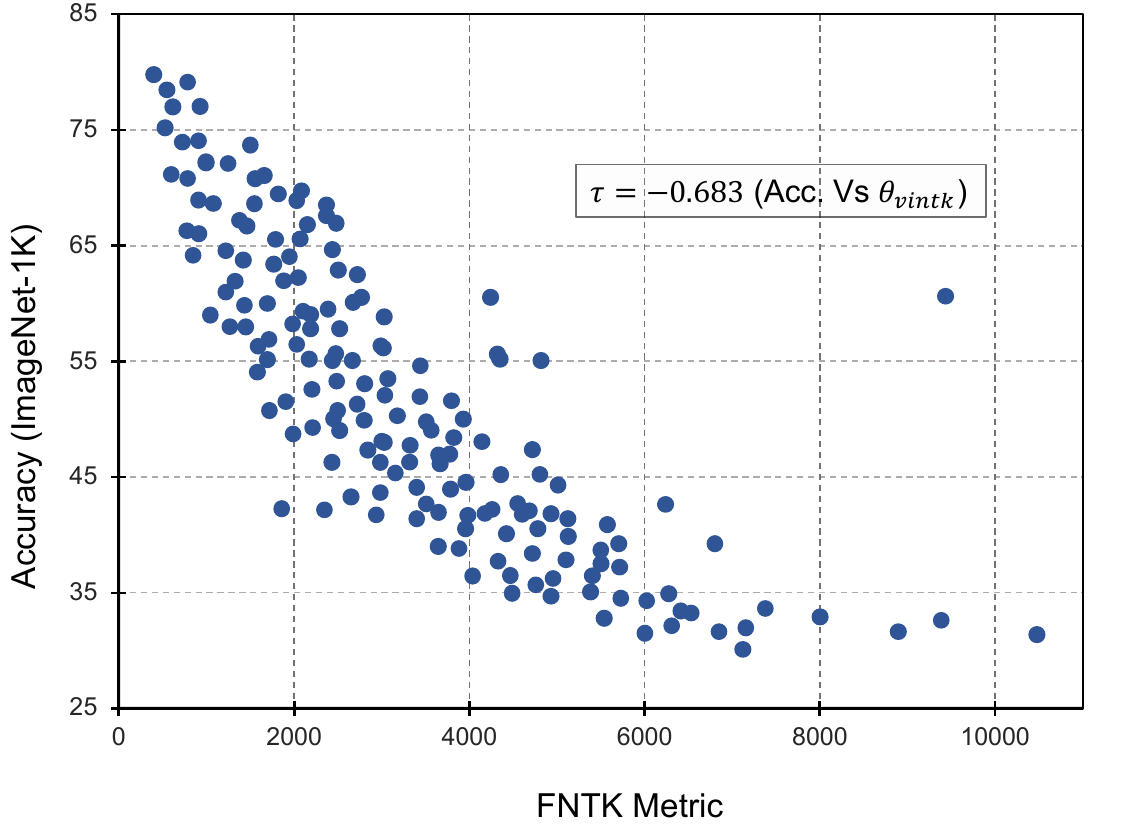}
     \caption{The correlation of our proposed ViNTK.}
     \label{fig:kendalltau}
\end{figure}

To verify the effectiveness of our proposed metric, we conduct a Kendall-Tau $\tau$ correlation test on ImageNet-1K, as shown in Figure~\ref{fig:kendalltau}. Note that all p-values for the tau correlation are less than 0.005 to ensure the results are statistically significant. We demonstrate a strong correlation between the accuracy and the ViNTK metric, showing the efficacy of our proposed metric in predicting ViT's performance. 

\begin{table}[t]
  \centering
  \caption{Image Classification on ImageNet-1k with NASViT search space. The model size range from 300M to 2G.}
  \label{tbl:nasvit_imagenet_comparison}
  \renewcommand*\arraystretch{1.1}
  \resizebox{0.4\textwidth}{!}{\begin{tabular}{c|l|c|c|c}
    \hline
    Size & Method & FLOPs (M) & Top-1 ($\%$) & Search Cost\\
    \hline
    \multirow{3}{*}{200M - 300M} & AlphaNet-A0~\cite{wang2021alphanet}  & 203 &  77.9  & 45 \\
    & NASViT-A0~\cite{gong2022nasvit} & 208 & 78.2 & 27 \\
    & \textbf{ViNTK-A0} & \textbf{210} & \textbf{78.3} & \textbf{1.0}\\
    \hline
    \multirow{4}{*}{300M - 400M} & LeViT~\cite{mehta2022mobilevit} & 300 & 76.6 & - \\
    & AlphaNet-A1~\cite{wang2021alphanet} & 317 &  79.4 & 75 \\
    & NASViT-A1~\cite{gong2022nasvit} & 309 & 79.7 & 27 \\
 & \textbf{ViNTK-A1} & \textbf{310} & \textbf{79.7} & \textbf{1.0}\\
    \hline
    \multirow{4}{*}{400M - 500M} & LeViT~\cite{mehta2022mobilevit} & 406 & 78.6 & - \\
    &  AlphaNet-A4~\cite{wang2021alphanet} & 444 & 80.4 & - \\
    & NASViT-A2~\cite{gong2022nasvit} & 421 & 80.5 & 27 \\
     & \textbf{ViNTK-A2} & \textbf{417} & \textbf{80.7} & \textbf{1.0}\\
    \hline

    \multirow{3}{*}{600M - 1G}& LeViT~\cite{mehta2022mobilevit} & 658 & 80.0 & - \\
    & NASViT-A5~\cite{gong2022nasvit} & 757 & 81.8 & 27 \\
    & \textbf{ViNTK-A5} & \textbf{772} & \textbf{82.0} & \textbf{1.0}\\
    \hline
    \multirow{3}{*}{1G - 2G} &
     PiT-XS~\cite{heo2021rethinking}  & 1,400 & 79.1 & -\\
    & NASViT-A6~\cite{gong2022nasvit} & 1,881 & 82.9 & 27 \\
    & \textbf{ViNTK-A6} & \textbf{1,910} & \textbf{82.9} & \textbf{1.0} \\
    \hline
  \end{tabular}}
\end{table}
\section{Experiments}
We validate the effectiveness of ImageNet-1K classification~\cite{krizhevsky2012imagenet} with two search spaces and semantic segmentation on both Cityscapes~\cite{cordts2016cityscapes} and ADE20K~\cite{zhou2017scene}.
\begin{figure}[t]
     \centering
     \includegraphics[width=0.43\textwidth]{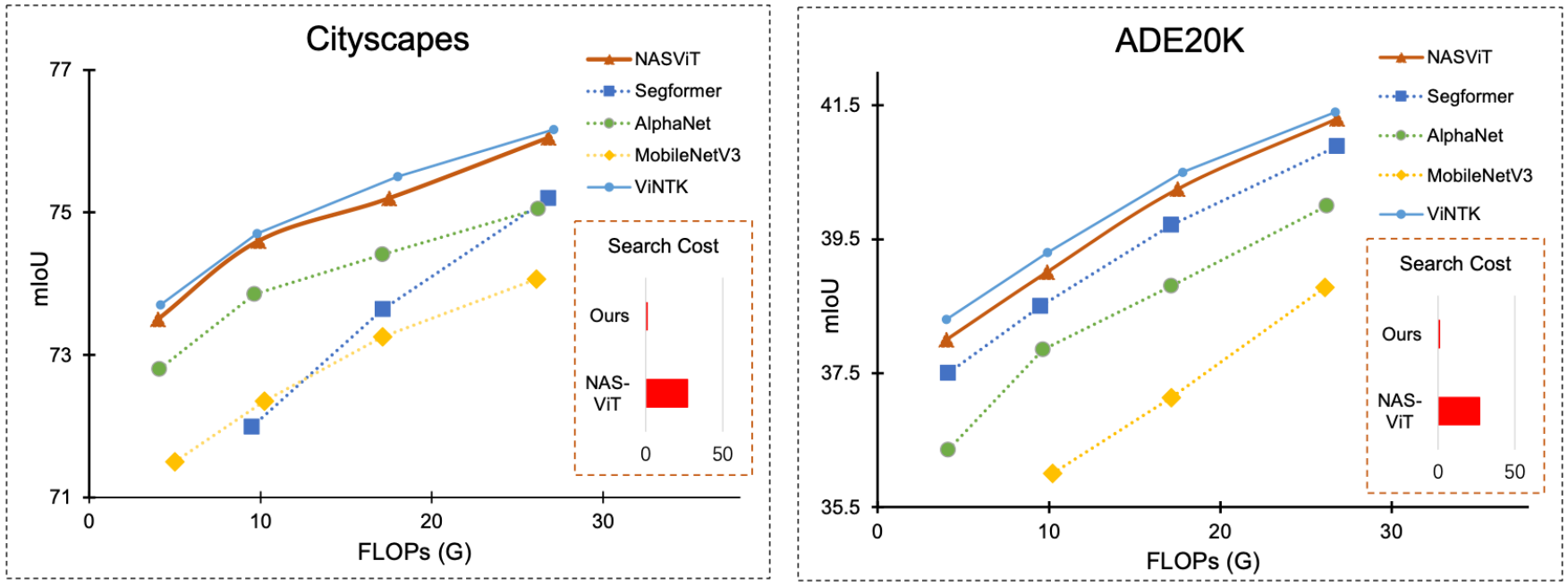}
     \caption{Experiments on semantics segmentation. }
     \label{fig:seg}
\end{figure}
\subsection{Image Classification on ImageNet-1K}

\noindent
\textbf{Searching Width and Depth on AutoFormer Search Space.} We tested our ViNTK in AutoFormer search space~\cite{chen2021autoformer}, comparing it to manual designs like DeiT~\cite{touvron2021training}, Swin~\cite{liu2021swin}, and NAS approaches such as ViTAS, AutoFormer, and TF-TAS. As shown in Table~\ref{tbl:autoformer}, ViNTK-identified architectures surpass manually-designed ViTs. Importantly, while our designs perform similarly to AutoFormer, our algorithm is 30x more efficient in search cost. Against TF-TAS, a training-free NAS, our approach either achieves better accuracy or reduced FLOPs, albeit at a slightly increased search cost.

\noindent
\textbf{Searching Width + Depth + Operations on NASViT Search Space.} We adopted NASViT search space for models ranging from 300M to 2G FLOPs, benchmarking against NASViT. As depicted in Table~\ref{tbl:nasvit_imagenet_comparison}, our method achieves similar or slightly superior results to NASViT but at 1/27th of the search cost. Traditional methods like NASViT and AlphaNet necessitate extended training on supernets, making them less suitable for resource-limited settings. Our approach proves to be significantly more efficient than leading NAS techniques.% tailored for ViT.

\subsection{Semantic Segmentation}
We further evaluate our method of semantic segmentation. Instead of directly finetuning our discovered architectures from ImageNet, we directly search the architecture via the target dataset. In particular, we search for four architectures with varying model sizes corresponding to the NASViT. On two benchmarks, Cityscapes~\cite{cordts2016cityscapes} and ADE20K~\cite{zhou2017scene}, we conduct separate search processes in order to find the optimal model. We set a hard threshold during our search process to ensure all architectures have similar FLOPs compared to NASViT. We also compare our method with SegFormer~\cite{xie2021segformer}, MobileNetv3~\cite{howard2019searching}, and AlphaNet~\cite{wang2021alphanet}. Similar to NASViT, we adopt the lightweight head from SegFormer as the decoder head for all backbones. Figure~\ref{fig:seg} presents the experimental results. By directly searching the target dataset, ViNTK achieved better performance than NASViT. For instance, on Cityscapes, among four different architecture, two of them achieve better performance than NASViT. As for ADE20K, all four data points obtain higher mIoU than NASViT. These results indicate that direct searching can be beneficial for training-free NAS.
\section{Conclusions}
Training-free NAS requires a deep understanding of deep neural networks. Our work thoroughly discusses the effectiveness of NTK-based training-free metrics in searching vision transformers from empirical study and theoretical analysis. To improve the inefficacy of NTK in approximating networks that learn high-frequency signals, we present a Fourier-NTK to resolve this issue. Moreover, we propose Fourier NTK to enhance the estimation power of training-free metrics. Our method significantly reduces the search cost of conventional training-based NAS for ViT.

\vfill\pagebreak
%\section{REFERENCES}
%\label{sec:refs}

\bibliographystyle{IEEEbib}
\bibliography{strings,refs}

\end{document}